\documentclass{article}

\usepackage[final]{neurips_2021}

\usepackage[utf8]{inputenc} 
\usepackage[T1]{fontenc}    
\usepackage{hyperref}       
\usepackage{url}            
\usepackage{booktabs}       
\usepackage{amsfonts}       
\usepackage{nicefrac}       
\usepackage{microtype}      
\usepackage{xcolor}         
\usepackage{graphbox}       
\usepackage{multirow}       
\usepackage{chemformula}    
\usepackage{makecell}       
\usepackage{tabularx}       
\usepackage{wrapfig}

\newcommand*\samethanks[1][\value{footnote}]{\footnotemark[#1]}

\newcommand\setrow[1]{\gdef\rowmac{#1}#1\ignorespaces}
\newcommand\clearrow{\global\let\rowmac\relax}
\clearrow

\title{ATOM3D: \\ Tasks On Molecules in Three Dimensions}

\author{%
  Raphael J. L. Townshend\thanks{Address correspondence to: \texttt{raphael@cs.stanford.edu}, \texttt{rondror@cs.stanford.edu}} \\
  Computer Science \\
  Stanford University \\
  \And
  Martin V{\"o}gele\thanks{Equal contribution.} \\
  Computer Science\\
  Stanford University \\
  \And
  Patricia Suriana\samethanks \\
  Computer Science \\
  Stanford University \\
  \And
  Alexander Derry\samethanks \\
  Biomedical Informatics \\
  Stanford University \\
  \And
  Alexander S. Powers \\
  Chemistry \\
  Stanford University \\
  \And
  Yianni Laloudakis \\
  Computer Science \\
  Stanford University \\
  \AND
  Sidhika Balachandar \\
  Computer Science \\
  Stanford University \\
  \And
  Bowen Jing \\
  Computer Science \\
  Stanford University \\
  \And
  Brandon Anderson \\
  Computer Science \\
  University of Chicago \\
  \And
  Stephan Eismann \\
  Applied Physics \\
  Stanford University \\
  \AND
  Risi Kondor \\
  Computer Science, Statistics \\
  University of Chicago \\
  \And
  Russ B. Altman \\
  Bioengineering, Genetics, Medicine \\
  Stanford University \\
  \And
  Ron O. Dror\samethanks[1] \\
  Computer Science \\
  Stanford University \\
}

\begin{document}

\maketitle

\begin{abstract}
Computational methods that operate on three-dimensional (3D) molecular structure have the potential to solve important problems in biology and chemistry. Deep neural networks have gained significant attention, but their widespread adoption in the biomolecular domain has been limited by a lack of either systematic performance benchmarks or a unified toolkit for interacting with 3D molecular data. To address this, we present ATOM3D, a collection of both novel and existing benchmark datasets spanning several key classes of biomolecules. We implement several types of 3D molecular learning methods for each of these tasks and show that they consistently improve performance relative to methods based on one- and two-dimensional representations. The choice of architecture proves to be important for performance, with 3D convolutional networks excelling at tasks involving complex geometries, graph networks performing well on systems requiring detailed positional information, and the more recently developed equivariant networks showing significant promise. Our results indicate that many molecular problems stand to gain from 3D molecular learning, and that there is potential for substantial further improvement on many tasks. To lower the barrier to entry and facilitate further developments in the field, we also provide a comprehensive suite of tools for dataset processing, model training, and evaluation in our open-source \texttt{atom3d} Python package. All datasets are available for download from \texttt{www.atom3d.ai}.
\end{abstract}

\newpage

\section{Introduction}
A molecule's three-dimensional (3D) shape is critical to understanding its physical mechanisms of action, and can be used to answer a number of questions relating to drug discovery, molecular design, and fundamental biology. While we can represent molecules using lower-dimensional representations such as linear sequences (1D) or chemical bond graphs (2D), considering the 3D positions of the component atoms---the atomistic geometry---allows for better modeling of 3D shape (Table \ref{tab:representation}). While previous benchmarking efforts, such as MoleculeNet \citep{Wu2018} and TAPE \citep{Rao2019}, have examined diverse molecular tasks, they focus on these lower-dimensional representations. In this work, we demonstrate the benefit yielded by learning on 3D atomistic geometry and promote the development of 3D molecular learning by providing a collection of datasets leveraging this representation.

Furthermore, the atom is emerging as a ``machine learning datatype'' in its own right, deserving focused study much like the pixels that make up images in computer vision or the characters that make up text in natural language processing. All molecules, including proteins, DNA, RNA, and drugs, can be represented as atoms in 3D space. These atoms can only belong to a fixed class of element types (carbon, nitrogen, oxygen, etc.), and all molecules are governed by the same underlying laws of physics that impose rotational, translational, and permutational symmetries. These systems also contain higher-level patterns that are poorly characterized, creating a ripe opportunity for learning them from data: though certain basic components are well understood (e.g. amino acids, nucleotides, functional groups), many others can not easily be defined. These patterns are in turn composed in a hierarchy that itself is only partially elucidated.

While deep learning methods such as graph neural networks (GNNs) and convolutional neural networks (CNNs) seem well suited to atomistic geometry, to date there has been no systematic evaluation of such methods on molecular tasks. Additionally, despite the growing number of 3D structures available in databases such as the Protein Data Bank (PDB) \citep{berman2000protein}, they require significant processing before they are useful for machine learning tasks. Inspired by the success of accessible databases such as ImageNet \citep{JiaDeng2009} and SQuAD \citep{Rajpurkar2016} in sparking progress in their respective fields, we create and curate benchmark datasets for atomistic tasks, process them into a simple and standardized format, systematically benchmark 3D molecular learning methods, and present a set of best practices for other machine learning researchers interested in entering the field of 3D molecular learning (see Section \ref{tips}). We reveal a number of insights related to 3D molecular learning, including the relative strengths and weaknesses of different methods and the identification of several tasks that provide great opportunities for 3D molecular learning. These are all integrated into the \texttt{atom3d} Python package to lower the barrier to entry and facilitate reproducible research in 3D molecular learning for machine learning practitioners and structural biologists alike.

\begin{table}
    
    \centering
    \caption{Representation choice for molecules. Adding in 3D information consistently improves performance. The depicted 1D representations are the amino acid sequences and SMILES strings \citep{Weininger1988} for proteins and small molecules, respectively.}
    \begin{tabular}{ cccc } 
        \toprule
        \multirow{2}{*}{\raisebox{-\heavyrulewidth}{Dimension}} &   \multirow{2}{*}{\raisebox{-\heavyrulewidth}{Representation}} &   \multicolumn{2}{c}{Examples} \\
        \cmidrule{3-4}
        & & Proteins & Small Molecules \\
        \midrule
        1D & linear sequence & KVKALPDA & CC(C)CC(C)NO
\\ 
        \midrule
        2D & chemical bond graph & \includegraphics[align=c]{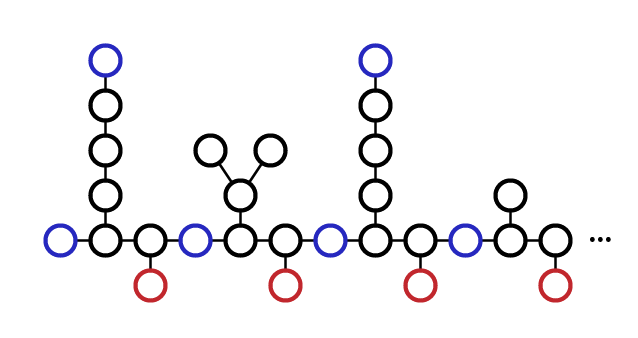} & \includegraphics[align=c]{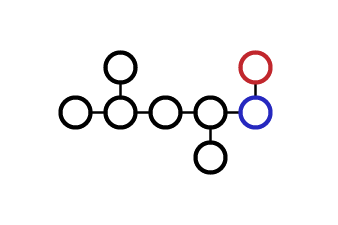} \\ 
        \midrule
        3D & atomistic geometry & \includegraphics[align=c]{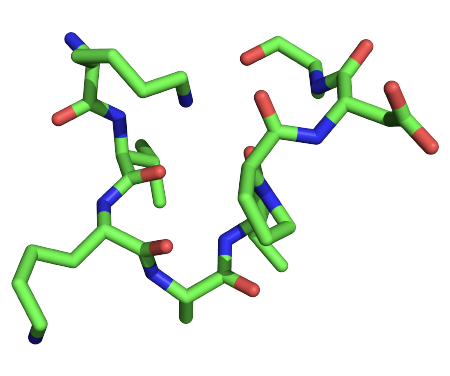} & \includegraphics[align=c]{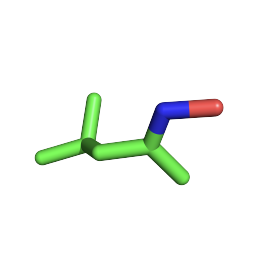} \\ 
        \bottomrule
    \end{tabular}
        
    \label{tab:representation}
\end{table}

\section{Related Work}

While three dimensional molecular data have long been pursued as an attractive source of information in molecular learning and chemoinformatics \citep{swamidass2005kernels, azencott2007one}, their utility has become increasingly clear in the last couple years.  Powered by increases in data availability and methodological advances, 3D molecular learning methods have demonstrated significant impact on tasks such as protein structure prediction \citep{senior2020improved,Jumper2021,Baek2021}, equilibrium state sampling \citep{Noe2019}, and RNA structure prediction \citep{Townshend2021}. At the same time, broader assessments of tasks involving molecular data have focused on either 1D or 2D representations \citep{Wu2018, Rao2019}. Through ATOM3D, we aim to provide a first benchmark for learning on 3D molecular data. There are a few major classes of algorithms that exist for data in this form.


Graph neural networks (GNNs) have grown to be a major area of study, providing a natural way of learning from data with complex spatial structure. Many GNN implementations have been motivated by applications to atomic systems, including molecular fingerprinting \citep{duvenaud2015convolutional}, property prediction \citep{Schuett2017,Gilmer2017,Liu2019-ngramgraph}, protein interface prediction \citep{Fout2017-nw}, and protein design \citep{Ingraham2019-us}. Instead of encoding points in Euclidean space, GNNs encode their pairwise connectivity, capturing a structured representation of atomistic data.  We note that some developed GNNs operate only on the chemical bond graph (i.e., 2D GNNs), with their edges representing covalent bonds, whereas others (including the ones we develop) operate on the 3D atomistic geometry (3D GNNs), with their edges representing distances between nearby pairs of atoms.

Three-dimensional CNNs (3DCNNs) have also become popular as a way to capture these complex 3D geometries. They have been applied to a number of biomolecular applications such as protein model quality assessment \citep{pages2019protein, derevyanko2018deep}, protein sequence design \citep{Anand2020-as}, protein interface prediction \citep{Townshend2019}, and structure-based drug discovery \citep{Wallach2015,Torng2017-da,Ragoza2017-yu, Jimenez2018-rl}. These 3DCNNs can encode translational and permutational symmetries, but incur significant computational expense and cannot capture rotational symmetries without data augmentation.

In an attempt to address many of the problems of representing atomistic geometries, equivariant neural networks (ENNs) have emerged as a new class of methods for learning from molecular systems. These networks are built such that geometric transformations of their inputs lead to well-defined transformations of their outputs. This setup leads to the neurons of the network learning rules that resemble physical interactions. Tensor field networks \citep{Thomas2018} and Cormorant \citep{Kondor2018,Anderson2019} have applied these principles to atomic systems and begun to demonstrate promise on larger systems such as proteins and RNA \citep{Eismann2020,Weiler2018-ef,Townshend2021}.

\section{Datasets for 3D Molecular Learning}
\label{datasets}

We select 3D molecular learning tasks from structural biophysics and medicinal chemistry that span a variety of molecule types. Several of these datasets are novel, while others are extracted from existing sources (Table \ref{tab:schematic}). We note that these datasets are intended for benchmarking machine learning representations, and do not always correspond to problem settings that would be seen in real-world scenarios. Below, we give a short description of each dataset's impact and source, as well as the metrics used to evaluate them and the splits. The splits were selected to minimize data leakage concerns and ensure generalizability and reproducibility. These datasets are all provided in a standardized format that requires no specialized libraries. Alongside these datasets, we present corresponding best practices (Appendix \ref{tips}) and further dataset-specific details (Appendix \ref{sup:datasets}).  Taken together, we hope these efforts will lower the barrier to entry for machine learning researchers interested in developing methods for 3D molecular learning and encourage rapid progress in the field.

\begin{table}[ht]
    \small
    \centering
     \caption{Tasks included in the ATOM3D datasets, along with schematic representations of their inputs. P indicates protein, SM indicates small molecule, R indicates RNA. Lines indicate interaction and a small square within a protein indicates an individual amino acid. New datasets are in bold.}
    \begin{tabular}{>{\rowmac}c>{\rowmac}c>{\rowmac}c>{\rowmac}c<{\clearrow} } 
        \toprule
        Name (Task Code) & Schematic & Objective & Source \\
        \midrule
            \makecell{Small Molecule \\ Properties (SMP)} &
            \includegraphics[align=c]{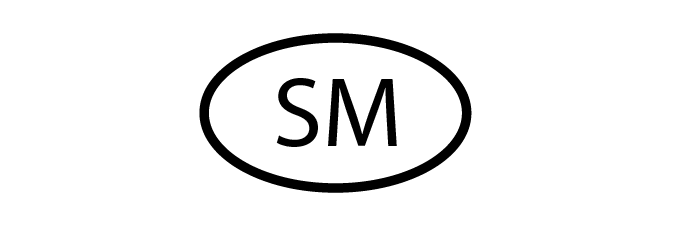} & 
            Properties & 
            \makecell{QM9 \\ \citep{Ruddigkeit2012}} \\ 
        \midrule
            \makecell{Protein Interface \\ Prediction (PIP)} &
            \includegraphics[align=c]{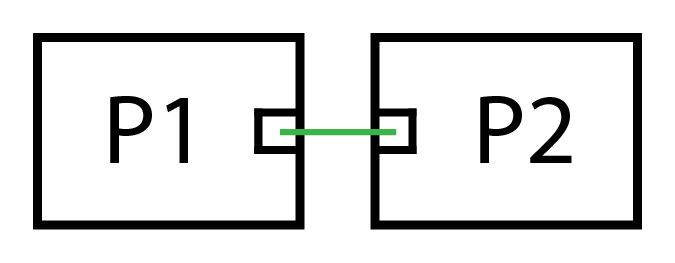} & 
            \makecell{Amino Acid \\ Interaction} &
            \makecell{DIPS \citep{Townshend2019} \\ DB5 \citep{Vreven2015}} \\ 
        \midrule
            \setrow{\bfseries}
            \makecell{Residue Identity \\ (RES)} & 
            \includegraphics[align=c]{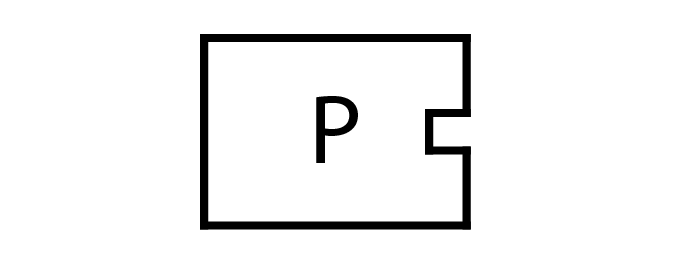} & 
            \makecell{Amino Acid \\ Identity} & 
            \makecell{New, created from PDB \\ \citep{berman2000protein}} \\ 
        \midrule
            \setrow{\bfseries}
            \makecell{Mutation Stability \\ Prediction (MSP)} &
            \includegraphics[align=c]{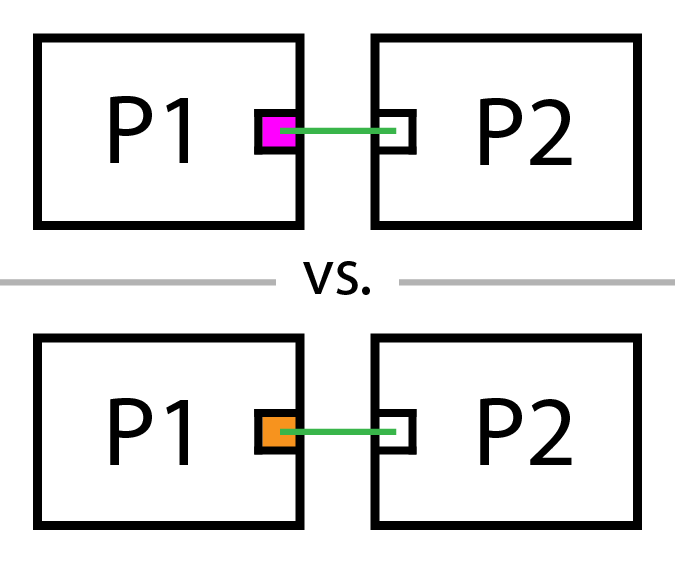} & 
            \makecell{Effect of \\ Mutation} & 
            \makecell{New, created from SKEMPI \\ \citep{jankauskaite2019skempi}} \\ 
        \midrule
            \makecell{Ligand Binding \\ Affinity (LBA)} &
            \includegraphics[align=c]{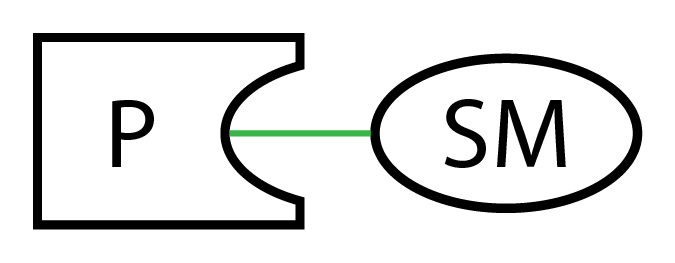} & 
            \makecell{Binding \\ Strength} & 
            \makecell{PDBBind \\ \citep{Wang2004}} \\ 
        \midrule
            \setrow{\bfseries}
            \makecell{Ligand Efficacy \\ Prediction (LEP)} &
            \includegraphics[align=c]{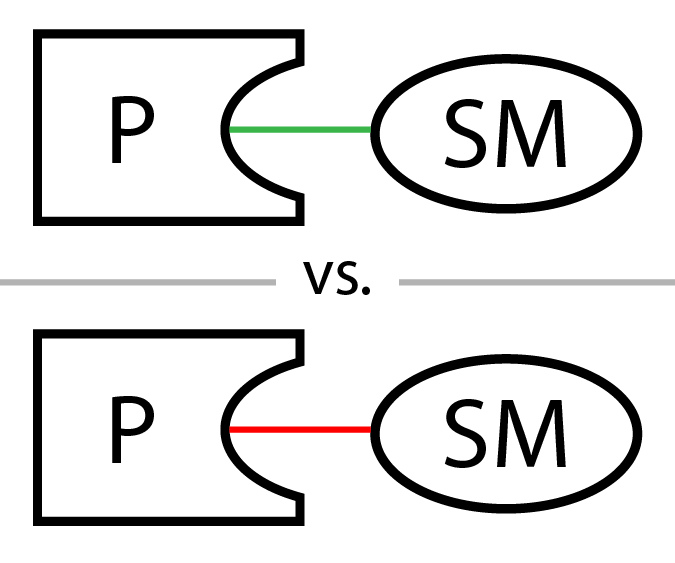} & 
            \makecell{Ligand \\ Efficacy} & 
            \makecell{New, created from PDB \\ \citep{berman2000protein}}  \\ 
        \midrule
            \makecell{Protein Structure \\ Ranking (PSR)} &
            \includegraphics[align=c]{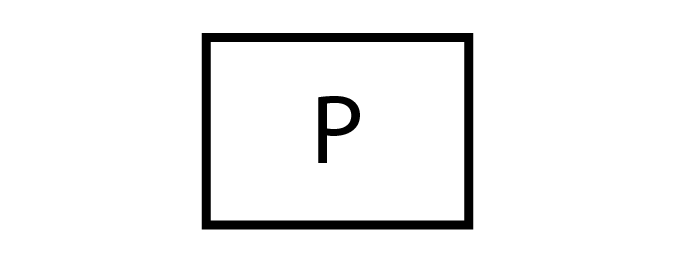} & 
            \makecell{Ranking} & 
            \makecell{CASP-QA \\ \citep{kryshtafovych2019critical}}  \\ 
        \midrule
            \makecell{RNA Structure \\ Ranking (RSR)} & 
            \includegraphics[align=c]{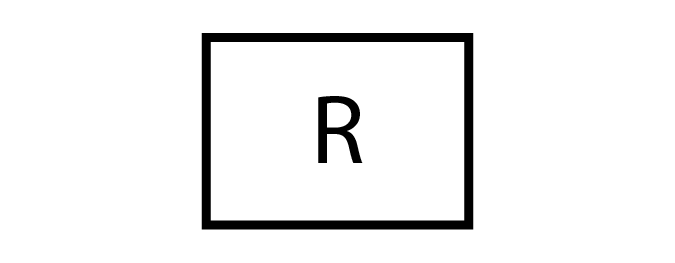} & 
            \makecell{Ranking} & 
            \makecell{FARFAR2-Puzzles \\ \citep{watkins2019farfar2}} \\
        \bottomrule
    \end{tabular}
       
    \label{tab:schematic}
\end{table}

\subsection{Small Molecule Properties (SMP)}
\textbf{Impact} -- Predicting physico-chemical properties of small molecules is a common task in medicinal chemistry and materials design. Quantum-chemical calculations can determine certain physico-chemical properties but are computationally expensive. \\
\textbf{Source} -- The QM9 dataset \citep{Ruddigkeit2012,Ramakrishnan2014} contains the results of quantum-chemical  calculations for 134,000 stable small organic molecules, each made up C, O, N, F, and H and including no more than nine non-hydrogen atoms. For each molecule, the dataset contains the calculated geometry of the ground-state conformation as well as calculated energetic, electronic, and thermodynamic properties. \\
\textbf{Targets} -- We predict the molecular properties from the ground-state structure. \\
\textbf{Split} -- We split molecules randomly.

\subsection{Protein Interface Prediction (PIP)}
\textbf{Impact} -- Proteins interact with each other in many scenarios---for example, antibody proteins recognize diseases by binding to antigens. A critical problem in understanding these interactions is to identify which amino acids of two given proteins will interact upon binding. \\
\textbf{Source} -- For training, we use the Database of Interacting Protein Structures (DIPS), a comprehensive dataset of protein complexes mined from the PDB \citep{Townshend2019}. We predict on the Docking Benchmark 5 \citep{Vreven2015}, a smaller gold standard dataset. \\
\textbf{Targets} -- We predict whether two amino acids will contact when their respective proteins bind. \\
\textbf{Split} -- We split protein complexes such that no protein in the training dataset has more than 30\% sequence identity with any protein in the DIPS test set or the DB5 dataset.

\subsection{Residue Identity (RES)}
\textbf{Impact} -- Understanding the structural role of individual amino acids is important for engineering new proteins. We can understand this role by predicting the propensity for different amino acids at a given protein site based on the surrounding structural environment \citep{Torng2017-da}. \\
\textbf{Source} -- We generate a novel dataset consisting of local atomic environments centered around individual residues extracted from non-redundant structures in the PDB.\\
\textbf{Targets} -- We formulate this as a classification task where we predict the identity of the amino acid in the center of the environment based on all other atoms. \\
\textbf{Split} -- We split environments by protein topology class according to the CATH 4.2 \citep{dawson2017cath}, such that all environments from proteins in the same class are in the same split dataset.

\subsection{Mutation Stability Prediction (MSP)}
\textbf{Impact} -- Identifying mutations that stabilize a protein's interactions is important to the design of new proteins. Experimental techniques for probing such mutations are labor-intensive \citep{antikainen2005altering, lefevre1997alanine}, motivating the development of efficient computational methods. \\
\textbf{Source} -- We derive a novel dataset by collecting single-point mutations from the SKEMPI database \citep{jankauskaite2019skempi} and model each mutation into the structure to produce a mutated structure. \\
\textbf{Targets} -- We formulate this as a binary classification task where we predict whether the stability of the complex increases as a result of the mutation. \\
\textbf{Split} -- We split protein complexes such that no protein in the test dataset has more than 30\% sequence identity with any protein in the training dataset.

\subsection{Ligand Binding Affinity (LBA)}
\textbf{Impact} -- Predicting the strength (affinity) of a candidate drug molecule's interaction with a target protein is a challenging but crucial task for drug discovery applications. \\
\textbf{Source} -- We use the PDBBind database \citep{Wang2004,Liu2015-op}, a curated database containing protein-ligand complexes from the PDB and their corresponding binding strengths (affinities). \\
\textbf{Targets} -- We predict $pK = -\log_{10}(K)$, where $K$ is the binding affinity in Molar units. \\
\textbf{Split} -- We split protein-ligand complexes such that no protein in the test dataset has more than 30\% sequence identity with any protein in the training dataset.

\subsection{Ligand Efficacy Prediction (LEP)}
\textbf{Impact} -- Many proteins switch on or off their function by changing shape. Predicting which shape a drug will favor is thus an important task in drug design. \\
\textbf{Source} -- We develop a novel dataset by curating proteins from several families with both "active" and "inactive" state structures, and model in 527 small molecules with known activating or inactivating function using the program Glide \citep{friesner2004glide}. \\
\textbf{Targets} -- We formulate this as a binary classification task where we predict whether a molecule bound to the structures will be an activator of the protein’s function or not. \\
\textbf{Split} -- We split complex pairs by protein target.

\subsection{Protein Structure Ranking (PSR)}
\textbf{Impact} -- Proteins are one of the primary workhorses of the cell, and knowing their structure is often critical to understanding (and engineering) their function. \\
\textbf{Source} -- We use the structural models submitted to the Critical Assessment of Structure Prediction (CASP) \citep{kryshtafovych2019critical}, a blind protein structure prediction competition, over the last 18 years.\\
\textbf{Targets} -- We formulate this as a regression task, where we predict the global distance test (GDT\_TS) of each structural model from the experimentally determined structure. \\
\textbf{Split} -- We split structures temporally by competition year.

\subsection{RNA Structure Ranking (RSR)}
\textbf{Impact} -- Similar to proteins, RNA plays major functional roles (e.g., gene regulation) and can adopt well-defined 3D shapes. Yet the problem is data-poor, with only a few hundred known structures.\\
\textbf{Source} -- We use the FARFAR2-Puzzles dataset, which consists of structural models generated by FARFAR2 \citep{watkins2019farfar2} for 20 RNAs from RNA Puzzles, a blind structure prediction competition for RNA \citep{cruz2012rna}. \\
\textbf{Targets} -- We predict the root-mean-squared deviation (RMSD) of each structural model from the experimentally determined structure. \\
\textbf{Split} -- We split structures temporally by competition year.

\section{Benchmarking Setup}
\label{setup}

To assess the benefits of 3D molecular learning, we use a combination of existing and novel 3D molecular learning methods, and implement a number of robust baselines. Our 3D molecular learning methods belong to one of each of the major classes of deep learning algorithms that have been applied to atomistic systems: graph networks, three-dimensional convolutional networks, and equivariant networks. Here we describe the main principles of the core networks used in these models. See Appendix \ref{sup:methods} for task-specific details and hyperparameters.

For GNNs, we represent molecular systems as graphs in which each node is an atom. Edges are defined between all atoms separated by less than $4.5$ Å, and weighted by the distance between the atoms using an edge weight defined by $w_{i,j} = \frac{1}{d_{i,j} + \epsilon}$, where $\epsilon = 10^{-5}$ is a small factor added for numerical stability. Node features are one-hot-encoded by atom type. Our core model uses five layers of graph convolutions as defined by \cite{kipf2016semi}, each followed by batch normalization and ReLU activation, and finally two fully-connected layers with dropout. For tasks requiring a single output for the entire molecular system, we use global pooling to aggregate over nodes. For tasks requiring predictions for single atoms or amino acids, we extract the relevant node embeddings from each graph after all convolutional layers (see Appendix \ref{sup:methods}).

For 3DCNNs, we represent our data as a cube of fixed size (different per task due to the different molecular sizes) in 3D space that is discretized into voxels with resolution of $1$ Å to form a grid (for PSR and RSR, we decrease the grid resolution to 1.3 Å in order to fit in the GPU memory). Each voxel is associated with a one-hot-encoded vector that denotes the presence or absence of each atom type. 
Our core model consists of four 3D-convolutional layers, each followed by ReLU activation and max-pooling (for every other convolution layer).  Two fully connected layers are applied after the convolutional layers to produce the final prediction. 

For ENNs, we use SE(3)-equivariant networks that represent each atom of a structure by its coordinates in 3D space and by a one-hot encoding of its atom type. No rotational augmentation is needed due to the rotational symmetry of the network. The core of all architectures in this work is Cormorant, a network of four layers of covariant neurons that use the Clebsch–Gordan transform as nonlinearity, as described and implemented by \cite{Anderson2019}.

\section{Benchmarking Results}

To assess the utility of 3D molecular learning, we evaluate our methods on the ATOM3D datasets and compare performance to state-of-the-art methods using 1D or 2D representations (for a comparison to the overall state-of-the-art, see Table \ref{tab:results_bp_sota}). We note that in many cases, 3D molecular learning methods have not been applied to the proposed tasks, and that several of the tasks are novel. In the following sections, we describe the results of our benchmarking and some key insights that can be derived from them. We also aggregate these results along with additional metrics and standard deviations over three replicates in Table \ref{tab:benchmarking-sup}. For each metric, we bold the best-performing method as well as those within one standard deviation of the best-performing method.  

\subsection{3D representations consistently improve performance}

Our evaluation of 3D methods on the tasks in ATOM3D reveals that incorporating atomistic geometry leads to consistently superior performance compared to 1D and 2D methods. For small molecules, state-of-the-art methods do not use 1D representations, so we focus instead on comparing to representations at the 2D level, i.e. the chemical bond graph. This is the approach taken by the 2D GNN introduced by \citep{Tsubaki2019} and the N-gram graph method by \citep{Liu2019-ngramgraph}, which both obtain similar results (Table~\ref{tab:results_sm}) on the small-molecule-only dataset SMP. When we add 3D coordinate information as in our ENN implementation, performance improves across all targets in SMP.

For tasks involving biopolymers (proteins and RNA), state-of-the-art methods do not use 2D representations, primarily because most of the chemical bond graph can be re-derived from the 1D representation, i.e. the linear sequence that makes up the biopolymer. We thus compare to representations at the 1D level (Table \ref{tab:results_bp}). For MSP and RES, both new datasets, we evaluate against the TAPE model \citep{Rao2019}, a transformer architecture that operates on protein sequence and is state-of-the-art amongst 1D methods for many tasks. For PIP, we compare to the sequence-only version of BIPSPI \citep{Sanchez-Garcia2018}, a state-of-the-art boosted decision tree method for protein interaction prediction. We find that 3D methods outperform these 1D methods on all biopolymer-only datasets (PIP, RES, MSP).

For tasks involving both biopolymers and small molecules, we compare to DeepDTA \citep{Ozturk2018a}. This network uses a 1D representation via a 1DCNN for both the biopolymer and small molecules. For LBA, we additionally compare to DeepAffinity \citep{karimi2019deepaffinity} which uses pairs of ligand SMILES strings and structurally annotated protein sequences.  Using a 3D representation for both the ligand and protein leads to improved or comparable performance for the joint protein--small molecule datasets (LBA and LEP, see Table \ref{tab:results_bpsm}).

\begin{table*}
    \small
    \centering
    \begin{minipage}[t]{4.5in}
    \caption{Small molecule results. Metric is mean absolute error (MAE).}
    \label{tab:results_sm}
    \end{minipage}
    \begin{tabular}{ cllccccc } 
        \toprule
        \multirow{2}{*}{Task} &  \multirow{2}{*}{Target} &   
        \multicolumn{3}{c}{3D} &
        \multicolumn{2}{c}{Non-3D} \\
        \cmidrule(lr){3-5} \cmidrule(lr){6-7}
      & & 3DCNN &  GNN & ENN & \citep{Tsubaki2019} & \citep{Liu2019-ngramgraph} \\ 
      \midrule
    \multirow{3}{*}{SMP} &   $\mu\,[\mathrm{D}]$ & 0.754 & 0.501 & \textbf{0.052} & 0.496 & 0.520 \\
          & $\varepsilon_\mathrm{gap}\, [\mathrm{eV}]$  & 0.580  &  0.137 & \textbf{0.095} & 0.154 & 0.184 \\
          & $U_0^\mathrm{at}\, [\mathrm{eV}]$ & 3.862 & 1.424 & \textbf{0.025} & 0.182 & 0.218 \\
                  \bottomrule
    \end{tabular}
\end{table*}

\begin{table*}
    \small
    \centering
    \begin{minipage}[t]{4.5in}
    \caption{Biopolymer results. AUROC is the area under the receiver operating characteristic curve. Asterisks  ($^{\ast}$) indicate that the exact training data differed (though splitting criteria were the same).}
    \label{tab:results_bp}
    \end{minipage}
    \begin{tabular}{ clccccc } 
    \toprule
    \multirow{2}{*}{Task} &   
    \multirow{2}{*}{Metric} &   
    \multicolumn{3}{c}{3D} &
    \multicolumn{1}{c}{Non-3D} \\
    \cmidrule(lr){3-5} \cmidrule(lr){6-6}
    
         & & 3DCNN &  GNN & ENN & \citep{Sanchez-Garcia2018} \\ 
        \midrule
        PIP & AUROC & \textbf{0.844} & $^{\ast}$0.669 & --- &  0.841 \\ 
        
        \toprule
        & & & & & \citep{Rao2019} \\
        \midrule
        RES & accuracy  & \textbf{0.451} & 0.082 & $^{\ast}$0.072 & $^{\ast}$0.30 \\
        \cmidrule(lr){1-6}
        MSP & AUROC & 0.574   & \textbf{0.609}  & 0.574 & 0.554 \\
        \bottomrule
    \end{tabular}
\end{table*}

\begin{table*}
    \small
    \centering
    \begin{minipage}[t]{4.5in}
    \caption{Joint small molecule/biopolymer results. $R_S$ is Spearman correlation, $R_P$ is Pearson correlation, AUROC is area under the receiver operating characteristic curve, and RMSE is root-mean-squared error.}
    \label{tab:results_bpsm}
    \end{minipage}
    \begin{tabular}{ clccccc } 
    \toprule
    \multirow{2}{*}{Task} &   \multirow{2}{*}{Metric} &   
    \multicolumn{3}{c}{3D} &
    \multicolumn{2}{c}{Non-3D} \\
    \cmidrule(lr){3-5} \cmidrule(lr){6-7}
      & & 3DCNN &  GNN & ENN & \citep{Ozturk2018a} & \citep{karimi2019deepaffinity} \\
    \midrule
    LBA & RMSE & \textbf{1.416} & 1.601 & 1.568 & 1.565 & 1.893\\ 
         & glob. $R_P$ & 0.550  & 0.545 & 0.389 & \textbf{0.573} & 0.415 \\ 
         & glob. $R_S$ & 0.553 & 0.533 & 0.408 & \textbf{0.574} & 0.426 \\  
        \midrule
    LEP & AUROC & 0.589 & \textbf{0.681} & \textbf{0.663} & \textbf{0.696} & --- \\
    \bottomrule
    \end{tabular}
\end{table*}

\begin{table*}
    \small
    \centering
    \begin{minipage}[t]{4.5in}
    \caption{Structure ranking results. $R_S$ is Spearman correlation. Mean measures the correlation for structures corresponding to the same biopolymer, whereas global measures the correlation  across all biopolymers.}
    \label{tab:results_sp}
    \end{minipage}
    \begin{tabular}{ clccccc } 
    \toprule
    \multirow{2}{*}{Task} &   
    \multirow{2}{*}{Metric} &   
    \multicolumn{3}{c}{3D} \\
    \cmidrule(lr){3-5}
      &  & 3DCNN &  GNN & SotA \\ 
      \midrule
      PSR
         & mean $R_S$ & \textbf{0.431} & 0.411 & \textbf{0.432} \citep{pages2019protein}\\
         &  glob. $R_S$ & \textbf{0.789} & 0.750 & \textbf{0.796} \citep{pages2019protein} \\ 
         \midrule
    RSR & mean $R_S$ & \textbf{0.264} & \textbf{0.234} & 0.173 \citep{watkins2019farfar2}  \\ 
         & glob. $R_S$ & 0.372 & \textbf{0.512} & 0.304 \citep{watkins2019farfar2}  \\ 
                  \bottomrule
    \end{tabular}
\end{table*}

The biopolymer structure ranking tasks (PSR and RSR) are inherently 3D in nature, as they involve evaluating the correctness of different 3D shapes taken on by the same biopolymer. Thus, critically, a 1D or 2D representation would not be able to differentiate between these different shapes since the linear sequence and chemical bond graph would remain the same. We therefore compare to state-of-the-art 3D methods as shown in Table \ref{tab:results_sp}, finding competitive or better results.  

More generally, we find that learning methods that leverage the 3D geometry of molecules hold state-of-the-art on the majority of tasks on our benchmark (Table \ref{tab:results_bp_sota}). These results demonstrate the potential of 3D molecular learning to address a wide range of problems involving molecular structure, and we anticipate that continued development of such models will aid progress in biological and chemical research.

\begin{wrapfigure}{R}{0.40\textwidth}
\includegraphics[width=0.39\textwidth]{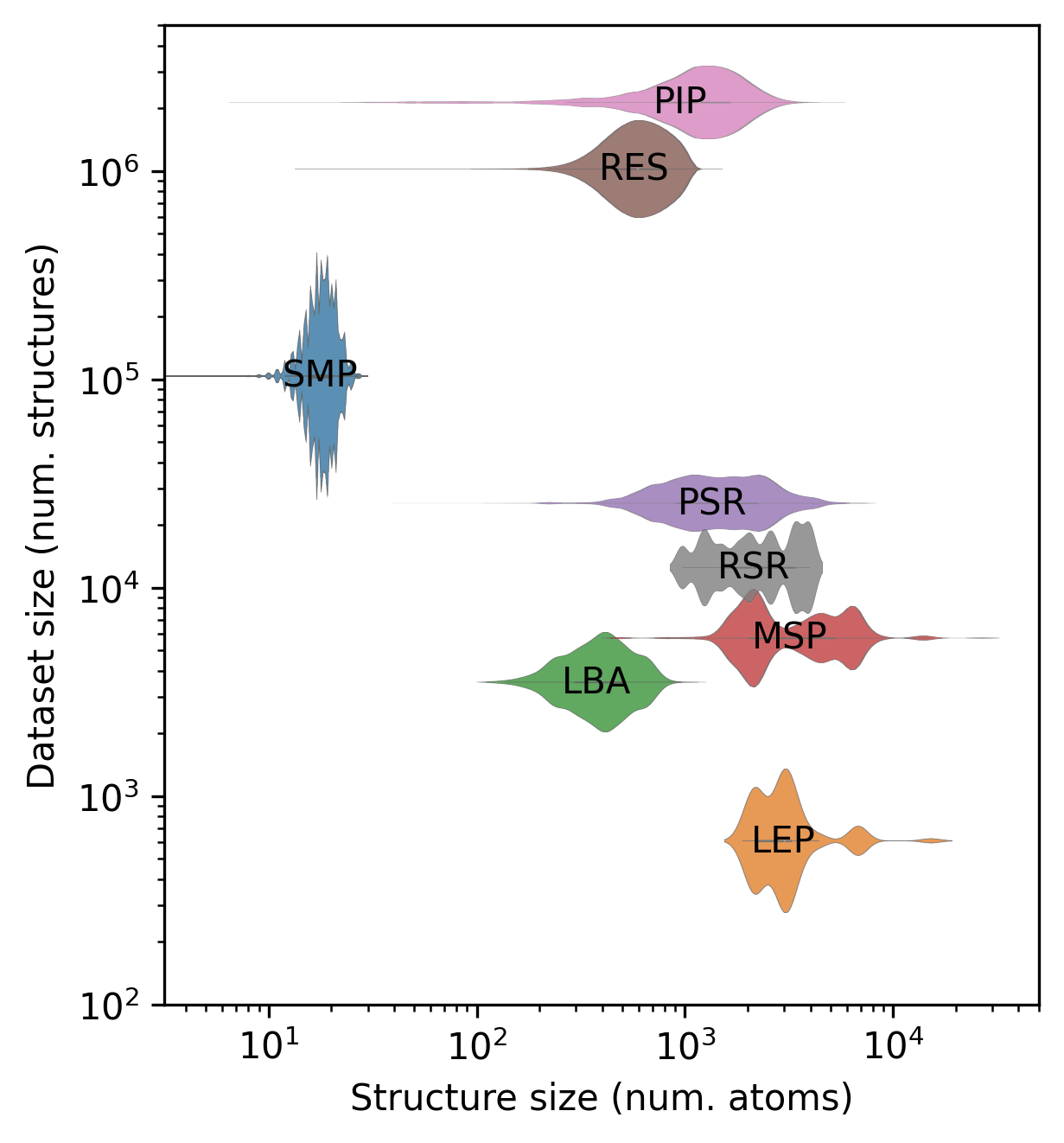}
\caption{Datasets plotted by their size, as well as the number of atoms in each one of their molecules.}
\label{fig:sizes}
\end{wrapfigure}

\subsection{Different tasks benefit from different architectures}


While 3D molecular learning methods outperform their non-3D counterparts and provide a systematic way of representing molecular data, our results also provide evidence that architecture selection plays an important role in performance.

For tasks focused on biopolymers and with large amounts of training data (PIP and RES, Figure \ref{fig:sizes}) we observe that 3DCNNs generally outperform standard GNNs. We hypothesize this is due to the ability of 3DCNNs to learn many-body patterns within a single filter, as opposed to GNNs that operate on one-body (node) and two-body (edge) features. Such many-body patterns are especially present in biopolymers, which generally adopt complex 3D geometries. This \textit{many-body representation} hypothesis implies that 3DCNNs have specific advantages in terms of representational power. 

However, as the size of the datasets decrease (Table \ref{tab:dataset_num_samples}), we see more even performance when comparing 3DCNNs and GNNs.  In particular, performance is quite similar on the intermediate-sized PSR and RSR datasets, and GNNs nearly fully supplant 3DCNNs on the small-sized MSP and LEP datasets. On these datasets, ENNs also equal or exceed 3DCNN performance. This is in line with the many-body representation hypothesis, as the increased representational power of 3DCNNs becomes less important in these data-poor regimes.


A notable exception to this trend is the large SMP dataset, where we see improved performance from GNNs and ENNs. We note, however, that this is the sole dataset involving only small molecules.  These molecules generally do not contain as complex of 3D geometries as biopolymers, and therefore do not contain large numbers of many-body patterns.  In addition, many of the prediction targets depend instead on the exact positions of each atom relative to its neighbors. While particle-based methods such as GNNs and ENNs can precisely record these distances, volumetric 3DCNNs must instead approximate these positions.  While increasing spatial resolution increases precision, it also leads to cubic scaling of complexity that prevents the same level of precision. 


Finally, equivariant networks show significant promise, despite being a recent innovation.
One motivation for their use is that they fill a ``happy medium'' where they both represent atom positions precisely and capture the many-body patterns present in complex geometries. On SMP,  the only dataset on which we tested ENNs without any limitations, we observed state-of-the-art performance. For other tasks, the performance of the ENN implementation we used limited us to training on a fraction of the data ($<$ 1\% for RES) or on a portion of the entire atomic structure (LBA, LEP, MSP), or did not permit us to apply it at all (PIP, PSR, RSR). Faster implementations are now available to allow scaling of ENNs to larger systems \citep{e3nn,gelib}.

\section{Conclusion}
\label{conclusion}

In this work we present a vision of the atom as a new ``machine learning datatype'' deserving focused study, as 3D molecular learning has the potential to address many unsolved problems in biology and chemistry. In particular, systems of atoms are well-suited to machine learning as they contain several underlying symmetries as well as poorly understood higher-level patterns. With ATOM3D, we take a first step towards this vision by providing a comprehensive suite of benchmark datasets and computational tools for building machine learning models for 3D molecular data.

We provide several benchmark datasets and compare the performance of different types of 3D molecular learning models across these tasks. We demonstrate that, for nearly all tasks that can be formulated in lower dimensions, 3D molecular learning yields gains in performance over 1D and 2D methods. We also show that selection of an appropriate architecture is critical for optimal performance on a given task; depending on the structure of the underlying data, a 3DCNN, GNN, or ENN may be most appropriate, especially in light of our many-body representation hypothesis. Equivariant networks in particular are continuing to improve in efficiency and stability, and we expect these to prove effective due to their ability to concisely model physical laws.

While this work demonstrates the potential of 3D structures and provides an initial set of benchmarks, there are some limitations to consider when using these resources. First, the datasets and tasks represented in ATOM3D are inherently biased towards biomolecules with solved structures. Certain classes of molecules (e.g. intrinsically disordered or transmembrane proteins) may therefore be underrepresented or absent, and the performance on these benchmarks will not necessarily generalize to such structures. Second, the benchmark models we report here are designed to be competitive but simple baselines. A bespoke architecture designed specifically for a certain task or molecule class and with comprehensive hyperparameter tuning is expected to outperform many of these baselines, and we encourage the exploration of novel and innovative approaches even within model classes that appear to underperform in these benchmarks (e.g. GNNs for the PIP task). 

Third, several of the benchmark tasks are formulated differently from those that biologists, chemists, and drug designers typically wish to solve. For example, when predicting the binding affinity of a ligand, one would rarely have access to a 3D structure of the ligand bound to the target (as in the LBA benchmark), because determining this structure experimentally would typically be far more expensive and time-consuming than measuring the ligand binding affinity. Likewise, when using machine learning methods to predict small-molecule properties more efficiently than through quantum chemical calculations (as in the SMP benchmark), one would not typically have access to the ground-state structure, because determining that structure requires equally expensive quantum chemical calculations. Although formulated in a somewhat artificial manner for convenience, such benchmarks have proven useful in evaluating general machine learning representations and methods. 

Finally, in addition to the datasets described here, there are many other open areas in biomedical research and molecular science that are ripe for 3D molecular learning, especially as structural data becomes readily available. Such tasks include virtual screening and pose prediction of small molecule drug candidates, as well as the incorporation of conformational ensembles instead of static structures in order to represent more faithfully the entire set of structures a molecule could adopt. Building on our easily extensible framework, we anticipate the addition of new datasets and tasks from across the research community.

Through this work, we hope to lower the entry barrier for machine learning practitioners, encourage the development of algorithms focused on 3D atomistic data, and promote an emerging paradigm within the fields of structural biology and medicinal chemistry.
\newpage

\section{Acknowledgments}
We thank Truong-Son Hy, Maria Karelina, David Liu, L{\'i}gia Melo, Joseph Paggi, and Erik Thiede for discussions and advice.
We also thank Aditi Krishnapriyan and Nicolas Swenson for pointing out an error in earlier GNN performance numbers.
This work was supported by the U.S. Department of Energy (DOE), Office of Science, Graduate Student Research (SCGSR) program (RJLT); EMBO Long-Term Fellowship ALTF 235-2019 (MV); an NSF Graduate Research Fellowship (PS); National Library of Medicine training grant LM012409 (AD); a Stanford Bio-X Bowes Fellowship (SE); NIH grants GM102365 and HG010615 (RBA); the Chan Zuckerberg Biohub (RBA); the DOE, Office of Science, Scientific Discovery through Advanced Computing (SciDAC) program (ROD); Intel (ROD); and DARPA Agreement No.~HR0011-18-9-0038 (RK). Most of the computing for this project was performed on the Sherlock cluster. We thank Stanford University and the Stanford Research Computing Facility for providing computational resources and support that contributed to these research results.

\bibliographystyle{plainnat}
\bibliography{reference}

\newpage

\newpage

\appendix

\section{Resource Availability and Licensing}

All datasets are available for download from \url{https://www.atom3d.ai}. The corresponding code for dataset processing and model training is maintained at \url{https://github.com/drorlab/atom3d}, along with instructions for installing our Python package. Further details can be found in the documentation at \url{atom3d.readthedocs.io/en/latest}. All datasets are hosted on Zenodo under individual DOIs, and are licensed under Creative Commons (CC) licenses. Full licenses, DOIs, and other details can be found on our website and Github. The authors bear all responsibility in case of violation of rights, and confirm the CC licenses for the included datasets.

\section{Broader Impact}
\label{sup:impact}
The methods and datasets presented here are intended to promote machine learning research that uses molecular structure. We expect that such research can be used in the development of new medicines and materials, as well as lead to a better understanding of human health. At the same time, the systems we propose are inherently complex, and thus can result in a greater degree of difficulty in interpreting their results, as well as preventing and assessing errors. These errors in turn can have severe consequences in both the development of new medicines, as well as in the treatment of patients using advances derived from this work. Careful evaluation of such systems results, as well as further research in increasing our ability to interpret their outputs, can help mitigate these concerns. Additionally, the benchmarks we use focus on molecules that have known structures. This is a biased set of all molecules, specifically in that they are molecules other researchers have chosen to focus on. Thus, we might expect trained methods to reflect these biases, and therefore have better performance in cases that are already being studied by the broader research community. This could lead to neglecting systems that are not as prioritized.

\section{Working with 3D Molecular Data Using ATOM3D}
\label{tips}

In order to facilitate the entry of new practitioners to the field of 3D molecular learning, we provide some high-level guidelines for working with the datasets we provide and for curating new ones. We provide computational tools required for these tasks in the \texttt{atom3d} package, and will be continuing to develop and support its functionality moving forward.

\subsection{Assembling New Datasets}

\paragraph{Data sources and repositories.}
The success of deep learning methods strongly depends on the availability of sufficient training data. Unless they have the capabilities to produce the necessary data, most scientists will rely on use public databases. The go-to repository for protein structures is the Protein Data Bank (PDB).\footnote{\url{https://www.rcsb.org/}} 
RNA structures can be found at the RNA 3D hub of Bowling Green State University.\footnote{\url{http://rna.bgsu.edu/rna3dhub/nrlist/} }
An exhaustive repository for small molecules is ChEMBL,\footnote{\url{https://www.ebi.ac.uk/chembl/}} though the 3D structures of small molecules are mostly not directly deposited. They can be generated by expensive quantum-chemical methods or in good approximation by cheminformatics tools such as RDKit. 
Many more specific databases are out there and worth being explored. \texttt{atom3d} provides methods to mine and convert data from many common formats in the field (e.g. PDB, SDF, XYZ) into our standardized dataset format.

\paragraph{Scope and limitations of the data.}
Even the most extensive databases cannot capture the large diversity of biological macromolecules or the space of potential drug molecules.
It is therefore necessary to think about the scientific problem at hand and whether the available data adequately represents the range of structures that are responsible for the studied effects.
An important general limitation of structural data is that molecules change conformation fluidly in real life due to thermal fluctuations and other effects. Additionally, interactions with other molecules, disordered regions, or environmental factors like pH can result in significant differences from their experimentally determined forms. 

\paragraph{Incomplete or corrupted data.}
Structural data is rarely perfect. Experimental uncertainties are mostly caused by limited resolution of the involved techniques such as X-ray crystallography or electron cryo-microscopy. Computationally generated structures are also prone to flaws in the underlying modeling programs (e.g., molecular force fields).

These limitations can lead to problems such as unrealistic conformations, missing or duplicate atoms, non-resolved amino acid side chains, and more. One has to decide whether to keep those structures or to sanitize them using computational tools. Additionally, hydrogen atoms are often not included in the data and, if needed for the task, have to be added when assembling the dataset. The most important guideline here is to be consistent and clear in the way these issues are treated. Sometimes it can be necessary to assemble two different datasets with different treatments of missing data. By providing a standardized format and functions for processing and filtering datasets through \texttt{atom3d}, we hope to simplify this process.

\subsection{Developing and Benchmarking New Algorithms}
\paragraph{Reading and preprocessing.} Algorithms represent data in various ways and a given dataset is not always compatible with the representation needed for the algorithm. For example, certain structures, residues, or atoms may need to be filtered out. Ideally, these steps are considered as dataset preparation and are separated from the algorithm itself, i.e. not hard-coded into the dataloader. This has two main advantages: (1) it saves time upon multiple reruns of the algorithm as structural data can be large and expensive to process, and (2) saving the preprocessed input dataset separately increases reproducibility, because small differences in preprocessing are often not recorded. We provide our benchmarking datasets in a format that is easy to read for most Python-based algorithms, and a simple interface for applying arbitrary transforms to convert between data formats (e.g. graph representations or voxelized grids).

\paragraph{Comparing algorithms.} Predictions can be tested with various metrics. Depending on the prediction problem, some of the metrics grasp the scientific aims of the training better than others. It is usually recommended to stick to the metrics that are common in the field and are given in the benchmarks. As science develops, new metrics for a specific problem might come up. These should be well justified and the old metrics should still be reported alongside them to allow for a comparison. Ideally, the new metrics are calculated for older models, too. To facilitate this in advance, when benchmarking an algorithm, specific predictions should be stored and not only metrics. In \texttt{atom3d}, we provide standardized evaluation classes to calculate the appropriate metrics from the outputs of a given algorithm.

\paragraph{Interpretation of results.}
When judging the performance of an algorithm, one should take into account the experimental uncertainties both in the structures but also in the label data. While small molecules can be investigated in much detail, it will rarely be possible to get near perfect performance for tasks involving complex biological macromolecules. 
Over time, even held-out test sets become part of the selection process for new methods as only those methods that perform better on the test set will prevail. A measured improvement can thus be caused by minor specifics of the test set. As the field matures and performance becomes saturated, the benchmark sets will still be valid as sanity checks for new methods, but harder tasks will be the ones driving new development. We anticipate that new tasks and datasets will be added to ATOM3D as the field evolves.


\section{Dataset Preparation}
\label{sup:datasets}

We present a set of methods to mine task-specific atomic datasets from several large databases (e.g. PDB) as well as to filter them, split them, and convert them to a format suitable for standard machine learning libraries (esp. PyTorch and TensorFlow). We store these datasets in LMDB format, where each atom is stored as a row in a standardized data frame. This data format accurately captures the natural hierarchy of atom subgroups in biomolecules, especially proteins, and enables data loading and processing to be consistent across datasets, tasks, and computational environments. The LMDB format also allows for labels and additional metadata is stored along with the atoms dataframe for each datapoint.

To capture hierarchical information in a way that is task-specific but standardized, we define an ``ensemble'' to be the highest-level of structure for each example, e.g. the PDB entry for the protein. Within each ensemble, we define ``subunits'', which represent the specific units of structure used for that task. For example, for the paired tasks (PIP, LEP, MSP), there is one subunit corresponding to each structure in the pair; for RES, there is one subunit for each residue microenvironment, and for structure ranking (PSR, RSR), there is one subunit for each candidate 3D structure. In this way, it is simple to iterate over each dataset and extract each atomistic structure, which can then be augmented and processed into any desired format (e.g. voxelized for the 3DCNN, converted to graphs for the GNN).

In the following sections, we describe the specific steps used to mine and process each dataset.

\subsection{Small Molecule Properties (SMP)}
The QM9 dataset is processed from the files provided on Figshare \citep{qm9}. The properties included in QM9 are the following:
\begin{itemize}
    \item $\mu$ - Dipole moment  (unit: $D$)
    \item $\alpha$ - Isotropic polarizability (unit: $\mathrm{bohr}^3$)
	\item $\epsilon_\mathrm{HOMO}$ - Highest occupied molecular orbital energy (unit: $\mathrm{Ha}$, reported in $\mathrm{eV}$)
	\item $\epsilon_\mathrm{LUMO}$ - Lowest unoccupied molecular orbital energy (unit: $\mathrm{Ha}$, reported in $\mathrm{eV}$)
	\item $\epsilon_\mathrm{gap}$ - Gap between HOMO and LUMO (unit: $\mathrm{Ha}$, reported in $\mathrm{eV}$)
	\item $R^2$ - Electronic spatial extent (unit: $\mathrm{bohr}^2$)
	\item $\mathrm{ZPVE}$ - Zero point vibrational energy (unit: $\mathrm{Ha}$, reported in $\mathrm{meV}$)
	\item $U_0$ - Internal energy at 0~K (unit: $\mathrm{Ha}$)
	\item $U_\mathrm{298}$ - Internal energy at 298.15~K (unit: $\mathrm{Ha}$)
	\item $H_\mathrm{298}$ - Enthalpy at 298.15~K (unit: $\mathrm{Ha}$)
	\item $G_\mathrm{298}$ - Free energy at 298.15~K (unit: $\mathrm{Ha}$)
	\item $C_v$ - Heat capacity at 298.15~K (unit: $\mathrm{\frac{cal}{mol\,K}}$)
\end{itemize}
It is common to subtract the reference thermochemical energy from $U_0$, $U_{298}$, $H_{298}$, $G_{298}$ to obtain:
\begin{itemize}
	\item $U_0^\mathrm{at}$ - Atomization energy at 0K (unit: $\mathrm{Ha}$, reported in $\mathrm{eV}$) 
	\item $U_\mathrm{298}^\mathrm{at}$ - Atomization energy at 298.15K (unit: $\mathrm{Ha}$, reported in $\mathrm{eV}$)
	\item $H_\mathrm{298}^\mathrm{at}$ - Atomization enthalpy at 298.15K (unit: $\mathrm{Ha}$, reported in $\mathrm{eV}$)
	\item $G_\mathrm{298}^\mathrm{at}$ - Atomization free energy at 298.15K (unit: $\mathrm{Ha}$, reported in $\mathrm{eV}$)
\end{itemize}
We report metrics for these quantities in the benchmark.

As recommended by the authors of the original dataset, we exclude 3,054 molecules that do not pass a geometrical consistency test \citep{Ramakrishnan2014}. Additionally, we excluded all 1,398 molecules that RDKit is unable to process - as in former GNN work \citep{Fey2019}. In this way, we ensure that all models in this work can be trained on the same data. Following previous work \citep{Wu2018,Gilmer2017,Schuett2017,Anderson2019}, we split the remaining dataset randomly in training, validation, and test set - containing 103547, 12943, and 12943 molecules, respectively.

\subsection{Protein Interface Prediction (PIP)}
For our test set, we download the cleaned PDB files from the DB5 dataset as provided in \citep{Townshend2019}, and convert to our standardized format. Each complex is an ensemble, with the bound/unbound ligand/receptor structures forming 4 distinct subunits of said ensemble. We use the bound forms of each complex to define neighboring amino acids (those with any heavy atoms within 6 Å of one another), and then map those onto the corresponding amino acids in the unbound forms of the complex (removing those that do not map). These neighbors are then included as the positive examples, with all other pairs being defined as negatives. At prediction time, we attempt to re-predict which possible pairings are positive or negative, downsampling negatives to achieve a 1:1 positive to negative split. We use the unbound subunits as our pair of input structures for testing. We use AUROC of these predictions as our metric to evaluate performance.

For our training set, we reproduce the Database of Interacting Protein Structures (DIPS) \citep{Townshend2019}. Specifically, we take the snapshot of all structures in the PDB from November 20, 2015. We apply a number of filtering operations, removing structures with no protein present, structures with less than 50 amino acids, structures with worse than 3.5 Å resolution, and structures not solved by X-ray crystallography or Cryo-EM. We then split the dataset into all pairs of interacting chains. These pairs form our ensembles, with each of the two chains being one subunit. We then remove pairs with less than 500 Å$^2$ buried surface area as measured by the FreeSASA Python library \citep{Mitternacht2016} (using total area computed the naccess classifier, including hydrogens and skipping unknown residues). Furthermore, to ensure there is no train/test contamination, we prune this set against the DB5 set defined above, removing any pairs that have a chain with more than 30\% sequence identity, using the software BLASTP \citep{Altschul1990}). We also prune the set based on structural similarity, removing any pairs in DIPS that map to corresponding SCOP \citep{Andreeva2014} pairs of superfamilies that are also present across a pair in DB5 (i.e., we remove a DIPS pair if the first subunit in that pair has a chain with a SCOP superfamily that is present in an unbound subunit of a DB5 pair, and the second subunit in that DIPS pair also has a SCOP superfamily that is present in the other unbound subunit of that same DB5 pair). Once this pruning is done, we split the DIPS set into a training, validation, and (internal) testing set based on PDB sequence clustering at a 30\% identity level, to ensure little contamination between them. We perform a 80\%, 10\%, 10\% split for training, validation, and testing, respectively. Note this internal testing set is not used for performance reporting. 

\subsection{Residue Identity (RES)}
Environments are extracted from a non-redundant subset of high-resolution structures from the PDB. Specifically, we use only X-ray structures with resolution $<$3.0 Å, and enforce a 60\% sequence identity threshold. We then split the dataset by structure based on domain-level CATH~4.2 topology classes \citep{dawson2017cath}, as described in \citep{Anand2020-as}. This resulted in a total of 21147, 964, and 3319 PDB structures for the train, validation, and test sets, respectively. Rather than train on every residue for each of these structures, we balance the classes in the train set by downsampling to the frequency of the least-common amino acid (cysteine). The original class balance is preserved in the test set. In total, the train, validation, and test sets comprised 3733710, 188530, and 1261342 environments, respectively. We ignore all non-standard residues. We represent the physico-chemical environment around each residue using all C, O, N, S, and P atoms in the protein and any co-crystallized ligands or ions. All non-backbone atoms of the target residue are removed, and each environment is centered around a ``virtual'' C$\beta$ position of the target residue defined using the average C$\beta$ position over the training set. 

\subsection{Mutation Stability Prediction (MSP)}
Mutation data are extracted from the SKEMPI 2.0 database \citep{jankauskaite2019skempi}. Non-point mutations or mutants that cause non-binding of the complex are screened out. Additionally, mutations involving a disulfide bond and mutants from the PDBs 1KBH or 1JCK are ignored due to processing difficulties. A label of 1 is assigned to a mutant if the $K_d$ of the mutant protein is less than that of the wild-type protein, indicating better binding, and 0 otherwise. Atoms from the twenty canonical amino acids were extracted from the PDBs provided in SKEMPI using PyMOL \citep{PyMOL}, and in silico mutagenesis is accomplished using PyRosetta \citep{chaudhury2010pyrosetta}, where dihedrals within 10 Å of the mutated residue are repacked. This protocol produces 893 positive examples and 3255 negative examples.
For ENN training, we use structures that are reduced to a size that is tractable for the implementation we used. To this end, we only selected the regions within a radius of 6~{\AA} around the C$\alpha$-atom of the mutated residue. For 3DCNNs, we analogously used a radius of 25~{\AA}. GNNs are trained on complete structures. This dataset is split by sequence identity at 30\%.

\subsection{Ligand Binding Affinity (LBA)}
PDBBind contains X-ray structures of proteins bound to small molecule and peptide ligands. We use the ``refined set'' (v.2019) consisting of 4,852 complexes filtered for various quality metrics, including resolution $\leq 2.5$ Å, R-factor $\leq 0.25$, lack of steric clashes or covalent bonding, and more \citep{Li2014-uz}. We further exclude complexes with invalid ligand bonding information. The binding affinity provided in PDBBind is experimentally determined and expressed in terms of inhibition constant ($K_i$) or dissociation constant ($K_d$), both in Molar units. As in previous works \citep{Ballester2010-wo, Zilian2013-ei, Ragoza2017-yu, Jimenez2018-rl}, we do not make the distinction between $K_i$ and $K_d$, and instead predict the negative log-transformed binding affinity, or $pK$. The majority of prior scoring functions have used the ``core set'' provided by the Critical Assessment of Scoring Functions (CASF) \citep{Su2019-wa} as a test set for evaluating prediction performance. However, by construction every protein in this test set is at least 90\% identical to several proteins in the training set. Thus, performance on this test set does not accurately represent the generalizability of a scoring function, and has been shown to overestimate the performance of machine learning models in particular \citep{Kramer2010-tf, Gabel2014-ah, li2017structural}. Therefore, to prevent overfitting to specific protein families, we create a new train/validation/test split based on a 30\% sequence identity threshold to limit homologous proteins appearing in both train and test sets. Specifically, we use a cluster-based approach to ensure that no protein in the training set has $>$ 30\% sequence identity to any protein in the validation or test sets, as calculated by BLASTP. To prevent overrepresentation of any single protein family (i.e. sequence identity cluster), we additionally enforce that no single cluster represents more than 20\% of the overall split. Splitting using this procedure resulted in training, validation, and test sets of size $3507$, $466$, and $490$, respectively. 

For comparison, we provide an additional, less restrictive, split based on a 60\% sequence identity threshold (results in Table~\ref{tab:benchmarking-sup}). This leads to training, validation, and test sets of size $3678$, $460$, and $460$, respectively.

For the ENN, we use a reduced dataset without hydrogens and only the most abundant heavy elements in the full dataset (C, N, O, S, Zn, Cl, F, P, Mg). From the binding pocket, we only use atoms within a distance of 6 Å from the ligand and only so many atoms as to not have more than 600 atoms in total (ligand + protein). This limitation of atom numbers is purely technical. The Kronecker products involved in the covariant neurons are memory intensive in the Cormorant implementation we used, and training on larger structures was limited by the memory of the GPUs available to us.

\subsection{Ligand Efficacy Prediction (LEP)}
Each input consists of a ligand bound to both the active and inactive conformation of a specific protein. The goal is to predict the label for this drug/ligand, either an “activator” or “inactivator” of the protein function. Why include these protein conformations in the input? From a biochemical perspective, if the drug binds much more favorably to the active protein conformation, it will act as an activator of the protein function. The model may then learn this differential binding strength to improve predictions of ligand function.

Pairs of structures for 27 proteins are obtained through manual curation of the Protein Data Bank structures where “active” and “inactive” conformational states are both available. For example, for ion channels, this means a channel in an open vs. closed state. 527 ligands with known protein binding and labeled function are selected from the IUPHAR database. We label ligands as activators if they are listed as “agonists” or “activators” and label ligands as inactivators if they are listed as “inhibitors” or “antagonists”. We select up to 15 of both activating and inactivating ligands for each protein.

We model the drugs bound to the relevant protein. To prepare protein structures for use in docking, we first prepare structures using the Schr{\"o}dinger suite. All waters are removed, the tautomeric state of the ligand present in the experimentally determined structure is assigned using Epik at pH 7.0 +/–2.0, hydrogen bonds are optimized, and energy minimization is performed with non-hydrogen atoms constrained to an RMSD of less than 0.3 Å from the initial structure. For ligands to be docked, the tautomeric state is assigned using Epik tool at target pH 7.0. Ligands are docked using default Glide SP. This results in 527 pairs of complexes. These are split into training, validation, and tests sets by protein target to ensure generalizability across proteins. 

For ENN training, we reduce the structures to a size that is tractable for the Cormorant implementation we used. To this end, we only use the regions within a radius of 5.5~{\AA} around the ligand. For 3D-CNNs, we use a radius of 25~{\AA}. GNNs were trained on complete structures. 

We require that efficacy predictions for a given ligand at a given protein not use information about efficacy of other ligands at that protein, to model a case when no such information is available. When efficacy measurements are available for other ligands at the same protein---as is the case for many well-studied drug targets---methods that take advantage of these (e.g., quantitative structure-activity relationship methods) may produce more accurate efficacy predictions.

\subsection{Protein Structure Ranking (PSR)}
The Critical Assessment of Structure Prediction (CASP) \citep{kryshtafovych2019critical} is a long-running international competition held biennially, of which CASP13 is the most recent, that addresses the protein structure prediction problem by withholding newly solved experimental structures (called \textit{targets}) and allowing computational groups to make predictions (called \textit{decoys}), which are then evaluated for their closeness to their targets after submission. Those submissions are then carefully curated and released as decoy sets in two stages (20 decoys per target for Stage 1, 150 decoys per target for Stage 2) for the Model Quality Assessment (MQA), one of the categories in CASP which aims to score a set of decoys of a target based on how closely they are to the target. For the PSR dataset, we compiled those decoys sets released in CASP5-13, then relaxed those structures with the SCWRL4 software \citep{Dunbrack2009} to improve side-chain conformations. For all decoys in the dataset, we computed the RMSD, TM-score, GDT\_TS, and GDT\_HA scores using the TM-score software \citep{tmscore}.

Mirroring the setup of the competition, we split the decoy sets based on target and released year. More specifically, we randomly split the targets in CASP5-10 and randomly sample 50 decoys for each target to generate the training and validation sets (508 targets for training, 56 targets for validation), and use the whole CASP11 Stage 2 as test set (85 targets total, with 150 decoys for each target). We chose CASP11 as test set, as the targets in CASP12-13 are not fully released yet.

\subsection{RNA Structure Ranking (RSR)}
The RNA Puzzles competition \citep{cruz2012rna} is a rolling international competition dealing with the RNA structure prediction problem. Similarly to CASP, newly solved experimental structures, referred to as natives, are withheld until computational groups make prediction, referred to as candidates. These candidates are then evaluated by their RMSD from the native. For this task, we use candidate structures created by the state-of-the-art structure generation method, FARFAR2 \citep{watkins2019farfar2}, for each of the 21 first RNA Puzzles. These are made available as part of the FARFAR2 publication. There are an average of 21303 (standard deviation of 13973) candidates generated per puzzles, with a large range of RMSDs. For the RSR dataset we randomly sample 1000 candidates per puzzle. We split temporally, by puzzle, using RNA Puzzles 1-13 for training, 14-17 (excluding 16) for validation, and 18-21 for testing.


\section{Task-Specific Experimental Details}
\label{sup:methods}

Below we describe the architectures and hyperparameters used for benchmarking. In general, these are intended to be robust but simple benchmarks for each task, so we did not undertake full tuning of every hyperparameter for every task, which would be very expensive. However, we did tune specific crucial hyperparameters such as learning rate, number of epochs, and 3DCNN grid size/resolution using a grid search methodology. Final models and hyperparameter settings were selected using performance on the validation set, with the held-out test sets only used to report final performance.



\subsection{3DCNNs}
Our base 3DCNN architecture consists of four 3D-convolutional layers with increasing filter size (32, 64, 128, and 256) --- each followed by ReLU activation, max-pooling (for every other convolution layer), and dropout --- and one fully-connected layer of size 512, followed by ReLU activation and dropout. For single model task (PSR, RSR, LBA, SMP), we add an additional fully-connected layer to transform to the required output dimension size. For paired tasks (PIP, LEP, MSP), we adapt this base architecture into a twin network, add an additional fully-connected layer followed by ReLU activation and dropout to combine the output of each member of the pair, and finally add a final fully-connected layer to transform to the required output dimension size, as in \citep{Townshend2019}.

For input to the 3DCNNs, we represent our data as cube in 3D space of certain radius ($40$ Å for PSR, RSR; $17$ Å for PIP; $20$ Å for LBA; $7.5$ Å for SMP; $25$ Å for LEP, MSP; $10$ Å for RES) that are discretized into voxels with resolution of $1$ Å to form a grid (for PSR and RSR, we need to decrease the resolution to 1.3 Å in order to fit them in the GPU memory). For paired tasks (PSR, RSR, and PIP), we form a separate voxel grid for each member of the pair. For most tasks, we use the centroid of each input structure as center of the grid, excluding LBA where we use the centroid of the ligand as center and MSP where we use the mutation point as center. Each grid voxel is associated with a binary feature vector which encodes the presence or absence of each specified atom type in that voxel. For PSR, RSR, PIP, and RES, we encode the presence of heavy atoms C, O, N, and S (P for RSR since S does not exist in RNA structures). For other tasks where hydrogen bonds might play an important role, we encode the hydrogen atom (H) in addition to C, O, N, and few other abundant atoms (F for LBA and SMP; S, Cl, F for LEP; S for MSP). To encode rotational symmetries, we apply a data augmentation strategy in which we apply 20 random rotations to the input grid, as in \citep{Townshend2019}, except for RES, where we instead apply the canonicalization procedure described in \citep{Anand2020-as}. 

For binary classification tasks, we use binary cross-entropy weighted by the class distribution (i.e. rarer class is weighted more heavily on mistakes). To address issues with imbalanced datasets, we randomly oversample/undersample the less/more frequent class respectively during training. For regression tasks, we use mean-squared error loss for training. All models were trained with Adam optimizer with default beta parameters and learning rate 0.0005 for SMP; 0.0001 for PSR, RSR, PIP, RES; 0.001 for LBA;  and 0.00001 for LEP, MSP. We monitor the loss on the validation set at every epoch. The weights of the best-performing are then used to evaluate on the held-out test set. The models were all trained on 1 Titan X(p) GPU  for 4--24 hours depending on the task. 

\subsection{GNNs}

Our base GNN architecture consists of five layers of graph convolutions as defined by Kipf and Welling \citep{kipf2016semi}, with increasing hidden dimension (64, 128, 128, 256, 256) each followed by batch normalization and ReLU activation. For tasks with paired input structures (PIP, LEP, MSP), we apply this convolutional architecture to each input structure separately in a twin network architecture with tied weights, and then concatenate the outputs before passing through two fully-connected layers of size 256 to transform to an output dimension of one neuron for binary classification. We regularize using dropout with a probability of 0.25 after the first fully-connected layer. Some tasks require classification of an entire structure, and thus are well-suited to graph-level outputs (PSR, RSR, LBA). Here, we apply global mean or addition pooling across all nodes before applying the final two layers. For PIP, RES, and MSP, instead of pooling we instead extract the embedding of the node corresponding to the C$\alpha$ atom of the residue in question (interacting residue, deleted residue, and mutated residue, respectively) after the final convolutional layer. For SMP, we use the previously-developed architecture presented in \citep{Gilmer2017}, which is publicly available.

We use a very simple featurization scheme for atomic systems. We define edges between all atoms separated by less than $4.5$ Å. Edges are weighted by the distance between the atoms, and nodes are featurized by one-hot-encoding all heavy (non-hydrogen) atoms. The only exception is SMP, where we adopt the established featurization scheme used in MoleculeNet \citep{Wu2018}. For tasks involving protein-ligand binding (LBA and LEP), we distinguish the protein and the ligand by using separate channels in the node features for each. All GNNs were implemented in PyTorch Geometric \citep{Fey2019}.

For binary tasks, we use a binary cross-entropy loss criterion weighted by the class distribution (e.g. a 1:4 positive:negative ratio would result in positive examples being up-weighted four-fold). For regression tasks, we use a mean-squared error criterion. For all models, we train with the Adam optimizer with learning rate 0.0001 (except for PIP, which uses a learning rate of 0.001) and monitor the relevant metrics (see Table \ref{tab:benchmarking-sup}) on the validation set after every epoch. The weights of the best-performing are then used to evaluate on the held-out test set. Models were all trained using 1 Tesla V100 GPU for 4--48 hours depending on the task.

Certain tasks involve making a prediction on a specific amino acid (PIP, RES, and MSP; see Table \ref{tab:schematic}), yet GNNs typically rely on summing over all node embeddings to compute a final graph embedding, making it difficult to isolate this amino acid. To remedy this, after our convolutional layers we extract the embedding of only the C$\alpha$ atom of the amino acid in question, thereby allowing our GNNs to isolate it.

\subsection{ENNs}

For the core of all Cormorant architectures in this work, we use a network of four layers of covariant neurons that use the Clebsch–Gordan transform as nonlinearity, with $L = 3$ as the largest index in the $SO(3)$ representation and 16 channels, followed by a single $SO(3)$-vector layer with $L = 0$. 
An input featurization network encodes the atom types as one-hot vectors.
For SMP, input and output are passed through multi-layer perceptrons (MLP) as in \citep{Anderson2019}. For the input, a weighted adjacency matrix with a learnable cut-off radius is constructed. This mask is passed alongside the input vector through a MLP with a single hidden layer with 256 neurons and ReLU activation. 
The output network is constructed from a set of scalar invariants that are passed through a network of two MLPs. Each of these MLPs has a single hidden layer of size 256, and the intermediate representation has 96 neurons.
For LBA and LEP, input and output layers are a single learnable mixing matrix, as used in the original Cormorant implementation for MD-17\citep{Anderson2019}.
The twin networks required for LEP and MSP was constructed by training two ENNs that are then connected by concatenating the single-network outputs which are then passed to a MLP analogous to the one described above for SMP. For MSP, the two structures corresponded to the wild-type structure and the mutated one; for LEP to the active and inactive one.
We extend the original Cormorant implementation to handle classification problems (binary and multi-class) and the twin network architecture. Our implementation\footnote{\url{https://github.com/drorlab/cormorant}} also allows to set a boundary on the Clebsch-Gordan product to eliminate training instabilities from a divergent loss that would otherwise arise occasionally for some of the tasks. 

We use MSE loss for regression tasks and cross-entropy loss for classification tasks. For all tasks, we used the AMSgrad optimizer with an initial learning rate of 0.001 and a final learning rate of 0.00001, decaying in a cosine function over the training process. We trained SMP and LBA for 150 epochs, LEP and MSP for 50 epochs, and RES for 30 epochs. We monitor the loss for the validation set after every epoch. The weights of the best-performing are then used to evaluate on the held-out test set. The models were all trained on 1 Titan X(p) GPU for 1--5 days depending on the task.


\section{1D and 2D Baselines}
\label{sup:baselines}

For each task, we select a baseline that fulfills the following criteria: (1) represents the current state-of-the-art (SOTA) for that task---or as close as possible---using only 1D (sequence only) or 2D (sequence and/or bond connectivity) molecular representations, and (2) either has a publicly available implementation or has reported results for the same task and splitting criteria. For PSR and RSR, which are inherently 3D tasks and have no appropriate 1D or 2D representation, we compare to the state-of-the-art 3D methods instead. Below we describe the choice and implementation of baseline models for each task.

\subsection{SMP}

As a 2D method for predicting molecular properties, we choose molecular GNNs \citep{Tsubaki2019} which are based on learning representations of subgraphs in molecules. We use an implementation that only uses the SMILES representation of the molecular graph.\footnote{\url{https://github.com/masashitsubaki/molecularGNN_smiles}}
As an additional 2D baseline, we compare to N-Gram Graph XGB \citep{Liu2019-ngramgraph}.
This method is based on an unsupervised representation called N-gram graph which first embeds the vertices in the molecule graph and then assembles the vertex embeddings in short walks in the graph. This representation is combined with the XGBoost learning method \citep{Chen2016-xgboost}.\footnote{\url{https://github.com/chao1224/n_gram_graph}}

\subsection{PIP}

For the PIP task, our non-3D method is the BIPSPI \citep{Sanchez-Garcia2018} model, a gradient-boosted decision tree. We compare to their model that uses only sequence and sequence conservation features and is evaluated on DB5.

\subsection{RES}

As a 1D sequence-based model for predicting residue identity, we choose the transformer architecture TAPE, introduced by \citep{Rao2019}. We use their reported accuracy on heldout families for language modeling, as that corresponds to a sequence-only version of our RES tasks, with similar stringency in terms of splitting criteria.

\subsection{MSP}

We use the publicly provided implementation of TAPE \citep{Rao2019}\footnote{\url{https://github.com/songlab-cal/tape}}. We modify their sequence-to-sequence head to predict the effect of mutations at specific positions, using the original unmutated protein as the input sequence and writing the output sequence as a one-hot-encoded 20-dimensional vector, indicating if a given mutation would be beneficial or detrimental. Note that the vast majority of positions would be unlabeled and therefore not included in the learning task.

\subsection{LBA}

As a 1D method for predicting ligand binding affinity, we choose DeepDTA \citep{Ozturk2018a}\footnote{\label{deepdta_github}\url{https://github.com/hkmztrk/DeepDTA}}, a 1DCNN based model that takes in pairs of ligand SMILES string and protein sequence as input. We use the same hyperparameters as in the original paper for the baseline.

We also compare our results against DeepAffinity \citep{karimi2019deepaffinity}\footnote{\label{deepaffinity_github}\url{https://github.com/Shen-Lab/DeepAffinity}}. We compare to their unified RNN/RNN-CNN model that takes in pairs of ligand SMILES string and their novel representations of structurally-annotated protein sequences (SPS/Structure Property-annotated Sequence) as input. Per the authors' recommendation, we use the DSSP software \citep{dssp1,dssp2} to generate the protein secondary structure and the protein relative solvent accessibility used in the SPS representation directly from the protein 3D structure, rather than the predicted ones by the SSpro/ACCpro software \citep{accpro1,accpro2} as done in the DeepAffinity paper. We use the same hyperparameters as in the original paper for the baseline, except for the maximum SMILES string and SPS lengths which we increase from 100 and 152 in the paper to 160 and 168, respectively, to accommodate for larger ligands/proteins in the PDBBind dataset. We used the pre-trained seq2seq encoders for proteins and ligands to initialize the joint supervised training of the encoders and CNN, and trained the DeepAffinity models for 1000 epochs. The pre-trained DeepAffinity seq2seq encoders were trained with maximum SMILES string and SPS lengths of 100 and 152, however, there are only 4\% of the ligands in the PDBBind dataset with SMILES string length larger than 100, and even much smaller percentage of the proteins (around 0.2\%) with SPS length larger than 152, so the input data distribution for PDBBind should still be in the range of that of DeepAffinity.

\subsection{LEP}

We train DeepDTA \citep{Ozturk2018a}\textsuperscript{\ref{deepdta_github}} (with the same hyperparameters as in the original paper) on the LEP dataset as baseline. As the inherent protein sequences and ligand SMILES strings are the same for both the inactive and active structures, the problem is reduced to binary classification task given a pair of protein sequence and the ligand SMILES string, and does not require modifying the DeepDTA architecture to make the twin network as in the GNN, ENN, or 3DCNN case. 

\subsection{PSR}

We compare our results against the state-of-the-art single-model methods as reported in \citep{pages2019protein}. These include 3DCNN \citep{3dcnn} and Ornate \citep{pages2019protein}, 3DCNN voxel-based methods trained on structural information, and Proq3D \citep{proq3d}, a deep-learning based method which employs structural information, Rosetta energy terms \citep{rosetta}, and evolutionary information derived from the amino acid sequence. We exclude ProteinGCN \citep{proteinGCN}, a recent GNN-based method, from comparison as they do not provide results on CASP11 dataset.

\subsection{RSR}

For RNA structure ranking, we compare our results against the Rosetta scoring function \citep{Alford2017}. In past RNA Puzzles competitions, methods using the Rosetta scoring function have been found to most consistently produce the lowest RMSD candidates. This is a physical- and knowledge-based potential specifically tuned for biomolecular structure.


\section{State-Of-The-Art Methods}
\label{sup:sota}

When possible, for tasks in Table \ref{tab:results_bp_sota}, we choose 3D methods that fulfill the following criteria: (1) they represent the current state-of-the-art for that task, or as close as possible, and (2) they either have a publicly available implementation or have reported results for the same task and splitting criteria. Here our choice of methods is described in more detail if not already discussed in the section above.

\subsection{SMP}
We compare to the state of the art, i.e., the best achieved prediction on each task, as reported in \cite{Anderson2019}. Many methods have been tested on QM9 and have reached excellent performance which makes them comparably hard to beat for new methods. In general, the best methods for QM9 so far are message passing neural networks \cite{Gilmer2017}, continuous-filter convolutional neural networks \cite{Schuett2017}, and Cormorant \cite{Anderson2019}. Differences in performance between earlier Cormorant studies and this work can be attributed to the different (random) split. 

\subsection{PIP}
We compare our results against the BIPSPI \cite{Sanchez-Garcia2018} model, a gradient-boosted decision tree. In contrast to the 1D/2D baseline comparison to BIPSPI, in this case we compare against their model that employs both structural- and sequence-based amino acid features.

\subsection{RES}

Since there have been no standardized datasets for this task to date, it is difficult to perform a direct comparison of methods. The closest comparison for a CNN trained on a balanced dataset of residue environments is 0.425, as reported in \cite{Torng2017-da}. While higher performance was reported by \cite{Anand2020-as} (accuracy 0.572), this model was trained on an unbalanced dataset comprising every standard residue environment in all training set PDBs. Similar performance has also been reported with other deep learning architectures \cite{Weiler2018-ef,Boomsma2017-mk}, but these do not describe their training/evaluation data or splitting criteria. In contrast, we restrict our training and evaluation to a balanced subset, downsampled to the frequency of the rarest class, which limits performance slightly. Additionally, to enable fair comparison over three replicates between 3DCNN and GNN, we then trained on only half of these down-sampled environments. The discrepancy in performance we observe on this subset is indicative of the fact that the differences in residue environment are subtle and complex, so simply increasing training data can result in higher performance. This is especially true for common classes such as leucine and glycine, which are over five times as frequent than the least common class, cysteine. Within these common classes, accuracy exceeds 80\%, which increases the average accuracy when classes are imbalanced.

\subsection{LEP}
Because this was a novel dataset, we computed initial results a non-deep learning method, Schr{\"o}dinger's Glide, to score each protein-ligand complex. Glide is state-of-the-art for scoring protein-ligand complexes and determining how "good" a pose is. This resulted in 2 scores per ligand; the score to the inactive protein structure and the score to the active protein structure. We then performed a binary classification by training an SVM on these two features to predict the ligand activity class. This approach is reasonable from a physical basis: if the ligand binds much better to the active protein structure than the inactive protein structure, then it will be an activator of the protein's function. 

\subsection{LBA}
Many methods have been developed for the prediction of ligand binding affinity using the PDBBind dataset. However, the standard has been to evaluate performance on the so-called ``core set'', as described in Section \ref{datasets}, after training and validating on the remainder of the refined set. The state-of-the-art reported on this core set has been achieved by the 3DCNN-based KDEEP \cite{Jimenez2018-rl}, followed closely by the popular random forest--based method RF-score \cite{Ballester2010-wo}. However, because the core set contains only proteins that are also present in the training set, this only measures \textit{in-distribution} performance, not generalizable scoring ability. Thus, the most comparable baseline for our dataset, which was split at 30\% sequence identity, is the performance of the empirical linear regression--based scoring function X-score fitted to complexes with less than 30\% identity to the core set, as reported in \cite{li2017structural}. We note that this is not a perfect comparison, since the procedure used in \cite{li2017structural} reduces the size of the training set significantly; however, as an empirical scoring function the performance of X-score is not very sensitive to training set size, compared to RF-score, which was significantly affected. 

\newpage

\section{Supplementary Tables}

\begin{table*}[h!]
    \small
    \centering
    \begin{minipage}[t]{4.5in}
    \caption{Comparison of performance against state-of-the-art methods, where available. The 3DCNN, GNN, and ENN networks achieve state-of-the-art in several tasks; for those where they do not (SMP, PIP, LBA), we note that the competing methods also use the 3D geometry of molecules.  Asterisks  ($^{\ast}$) indicate that the exact training data differed (though splitting criteria were the same).}
    \vspace{0.25cm}
    \label{tab:results_bp_sota}
    \end{minipage}
    \begin{tabular}{ clcccrr } 
    \toprule
    \multirow{2}{*}{Task} &   
    \multirow{2}{*}{Metric} &   
    \multicolumn{3}{c}{3D} &
    \multicolumn{2}{c}{SOTA} \\
    \cmidrule(lr){3-5} 
        
         & & 3DCNN &  GNN & ENN  \\ 
        \midrule
        SMP &   $\mu\,[\mathrm{D}]$ & 0.754 & 0.501 & 0.052 & \textbf{$^{\ast}$0.030} & \citep{Gilmer2017} \\
          & $\varepsilon_\mathrm{gap}\, [\mathrm{eV}]$  & 0.580  &  0.137 & 0.095 & \textbf{$^{\ast}$0.061} & \citep{Anderson2019} \\
          & $U_0^\mathrm{at}\, [\mathrm{eV}]$ & 3.862 & 1.424 & 0.025 & \textbf{$^{\ast}$0.014} & \citep{Schuett2017} \\
          \midrule
        PIP & AUROC & 0.844 & $^{\ast}$0.669 & --- & \textbf{0.919} & \citep{Sanchez-Garcia2018} \\ 
       \midrule
        RES & accuracy  & \textbf{0.451} & 0.082 & $^{\ast}$0.072 & $^{\ast}$0.425 & \citep{Torng2017-da} \\
        \midrule
        MSP & AUROC & 0.574   & \textbf{0.609}  &  0.574 & --- \\
        \midrule
        LBA & RMSE & \textbf{1.416} & 1.601 & 1.568 & $^{\ast}$1.838 & \citep{li2017structural} \\ 
         & glob. $R_P$ & 0.550  & 0.545 & 0.389 & \textbf{$^{\ast}$0.645} & \citep{li2017structural}\\ 
         & glob. $R_S$ & 0.553 & 0.533 & 0.408 & \textbf{$^{\ast}$0.697} & \citep{li2017structural} \\ 
        \midrule
        LEP & AUROC & 0.589 & 0.681 & 0.663 & \textbf{0.770} & \citep{friesner2004glide}  \\
        \midrule
        PSR
         & mean $R_S$ & \textbf{0.431} & 0.411 & --- & \textbf{0.432} & \citep{pages2019protein}\\
         &  glob. $R_S$ & \textbf{0.789} & 0.750 & --- & \textbf{0.796} & \citep{pages2019protein} \\ 
         \midrule
         RSR & mean $R_S$ & \textbf{0.264} & \textbf{0.234} & --- & 0.173 & \citep{Alford2017}  \\ 
         & glob. $R_S$ & 0.372 & \textbf{0.512} & --- & 0.304 & \citep{Alford2017}  \\ 
        \bottomrule
    \end{tabular}
\end{table*}

\begin{table*}[h!]
    \footnotesize
    \centering
    \caption{Complete benchmarking results from Tables \ref{tab:results_sm}--\ref{tab:results_sp}, with additional metrics and standard deviations reported over three replicates. $R_K$ is Kendall correlation and AUPRC is area under the precision-recall curve.  SMP metrics are all mean absolute error (MAE). Asterisks  ($^{\ast}$) indicate that the exact training data differed (though splitting criteria were the same).}
    \label{tab:benchmarking-sup}
    \begin{tabular}{ llrrrr } 
    \toprule
     Task& Metric & 3DCNN &  GNN & ENN & SOTA Baseline \\
     &  & & & &\citep{pages2019protein} \\ 
      \midrule
      PSR
          & mean $R_P$ & \textbf{0.557 $\pm$ 0.011} & 0.500 $\pm$ 0.012 & --- & 0.444 \\
               &  mean $R_K$ & \textbf{0.308 $\pm$ 0.010} & 0.289 $\pm$ 0.005 & --- & 0.304  \\
               &  mean $R_S$ & \textbf{0.431 $\pm$ 0.013} & 0.411 $\pm$ 0.006 & --- & \textbf{0.432}  \\
               &  global $R_P$ & \textbf{0.780 $\pm$ 0.016} &  0.747 $\pm$ 0.018 & --- & 0.772  \\
               &  global $R_K$ & \textbf{0.592 $\pm$ 0.016} & 0.547 $\pm$ 0.016 & --- & \textbf{0.594}  \\
               &  global $R_S$ & \textbf{0.789 $\pm$ 0.017} & 0.750 $\pm$ 0.018 & --- & \textbf{0.796}  \\ 
         
         \midrule
      &  &  &   &  & SOTA Baseline \\
     &  & & & &\citep{Alford2017} \\ 
      \midrule
         
    RSR  & mean $R_P$ & \textbf{0.286 $\pm$ 0.038} & \textbf{0.275 $\pm$  0.007} & --- & 0.129 \\
                   & mean $R_K$ & \textbf{0.181 $\pm$ 0.032} &  \textbf{0.157 $\pm$  0.004} & --- & 0.119  \\ 
                   & mean $R_S$ & \textbf{0.264 $\pm$ 0.046} & \textbf{0.234 $\pm$  0.006} & --- & 0.173 \\ 
                   & global $R_P$ & 0.360 $\pm$ 0.030 & \textbf{0.519 $\pm$ 0.051} & --- & 0.161 \\ 
                   & global $R_K$ & 0.247 $\pm$ 0.017 & \textbf{0.348 $\pm$  0.038} & --- & 0.206 \\
                   & global $R_S$ & 0.372 $\pm$ 0.027 & \textbf{0.512 $\pm$  0.049} & --- & 0.304  \\ 
         
         \midrule
       &  &  &   &  & Non-3D Baseline \\
     &  & & &  \multicolumn{2}{r}{\citep{Sanchez-Garcia2018}} \\ 
      \midrule
        
    PIP & AUROC & \textbf{0.844 $\pm$ 0.002} & $^{\ast}$0.669 $\pm$ 0.001 & --- &  0.841 \\ 
    \midrule
       &  &  &   &  & Non-3D Baseline \\
     &  & & & &\citep{Rao2019} \\ 
      \midrule
   RES  & accuracy  & \textbf{0.451 $\pm$ 0.002} & 0.082 $\pm$ 0.002 & $^{\ast}$0.072 $\pm$ 0.005 & *0.30  \\
   MSP  & AUROC     & 0.574 $\pm$ 0.005  & \textbf{0.609 $\pm$ 0.011} & 0.574 $\pm$ 0.040 & 0.554  \\
        & AUPRC     & 0.187 $\pm$ 0.007 & 0.176 $\pm$ 0.003  & \textbf{0.196 $\pm$  0.010} &  --- \\ 
       
       \midrule
      &  &  &   &  & Non-3D Baseline \\
     &  & & & &\citep{Tsubaki2019} \\ 
      \midrule
    SMP & $\mu\,[\mathrm{D}]$ & 0.754 $\pm$ 0.009 & 0.501 $\pm$  0.002 & \textbf{0.052 $\pm$  0.007} & 0.496 $\pm$ 0.002 \\
        & $\alpha\, [\mathrm{bohr}^3$] & 3.045 $\pm$ 1.128 & 1.562 $\pm$  0.038 & \textbf{0.127 $\pm$  0.026} & 0.392 $\pm$ 0.004\\
        & $\varepsilon_\mathrm{HOMO}\, [\mathrm{eV}]$   & 0.303 $\pm$ 0.000 & 0.092 $\pm$  0.002 & \textbf{0.044 $\pm$ 0.010} & 0.107 $\pm$ 0.001\\
        & $\varepsilon_\mathrm{LUMO}\, [\mathrm{eV}]$   & 0.517 $\pm$ 0.011 & 0.096 $\pm$  0.001 & \textbf{0.035 $\pm$ 0.003} & 0.115 $\pm$ 0.001\\
        & $\varepsilon_\mathrm{gap}\, [\mathrm{eV}]$   & 0.580  $\pm$ 0.004 &  0.137 $\pm$  0.002 & \textbf{0.095 $\pm$  0.021} & 0.154 $\pm$ 0.001\\
        & $R^2\, [\mathrm{bohr}^2]$  & 64.514 $\pm$ 1.524 &  89.912 $\pm$ 32.591 & \textbf{1.045 $\pm$  0.065} & 27.976 $\pm$ 0.212\\
        & $\mathrm{ZPVE}\, [\mathrm{meV}]$ & 88.219 $\pm$ 16.287 &  33.504 $\pm$  5.548 & \textbf{1.705 $\pm$ 0.044} & 10.614 $\pm$ 0.270 \\
        & $U_0^\mathrm{at}\, [\mathrm{eV}]$ & 3.862 $\pm$ 0.594 & 1.424 $\pm$  0.211 & \textbf{0.025 $\pm$  0.001} & 0.182 $\pm$ 0.004 \\
        & $U_{298}^\mathrm{at}\, [\mathrm{eV}]$  & 4.356 $\pm$ 0.498 &   1.227 $\pm$  0.610 & \textbf{0.025 $\pm$  0.001} & 0.181 $\pm$ 0.001 \\
        & $H_{298}^\mathrm{at}\, [\mathrm{eV}]$  & 4.088 $\pm$ 0.229 &  0.927 $\pm$  0.177 & \textbf{0.024 $\pm$ 0.001} & 0.180 $\pm$ 0.004 \\
        & $G_{298}^\mathrm{at}\, [\mathrm{eV}]$  & 4.369 $\pm$ 0.805 &  1.171 $\pm$  0.497 & \textbf{0.024 $\pm$ 0.001} & 0.173 $\pm$ 0.000\\
        & $C_v$\, $[\frac{\mathrm{cal}}{\mathrm{mol K}}]$ & 1.418 $\pm$ 0.200 & 0.350 $\pm$  0.078 & \textbf{0.034 $\pm$ 0.001} & 0.187 $\pm$ 0.004\\
        \midrule
      &  &  &   &  & Non-3D Baseline \\
     &  & & & &\citep{Ozturk2018a} \\ 
      \midrule
        
    LBA  & RMSE & \textbf{1.416 $\pm$ 0.021} & 1.601 $\pm$  0.048 & 1.568 $\pm$ 0.012 & 1.565 $\pm$ 0.018 \\
    (30\%)              & global $R_P$ & 0.550 $\pm$ 0.021 & 0.545 $\pm$  0.027 & 0.389 $\pm$ 0.024 & \textbf{0.573 $\pm$ 0.022} \\ 
                 & global $R_S$ & 0.553 $\pm$ 0.009 & 0.533 $\pm$ 0.033 & 0.408 $\pm$ 0.021 & \textbf{0.574 $\pm$ 0.010} \\  
    LBA  & RMSE & 1.621 $\pm$ 0.025 & \textbf{1.408 $\pm$ 0.069} & 1.620 $\pm$ 0.049 & 1.760 $\pm$ 0.415 \\ 
         (60\%)          & global $R_P$ & 0.608 $\pm$ 0.020 & \textbf{0.743 $\pm$  0.022} & 0.623 $\pm$ 0.015 & 0.713 $\pm$ 0.013 \\
                   & global $R_S$ & 0.615 $\pm$ 0.028 & \textbf{0.743 $\pm$ 0.027} & 0.633 $\pm$  0.021 & 0.702 $\pm$ 0.013 \\ 
    
    LEP &  AUROC & 0.589 $\pm$ 0.020 & \textbf{0.681 $\pm$ 0.062} & \textbf{0.663 $\pm$ 0.100} & \textbf{0.696 $\pm$ 0.021}  \\
          & AUPRC & 0.483 $\pm$ 0.037 & \textbf{0.598 $\pm$ 0.135} & \textbf{0.551 $\pm$ 0.121} & \textbf{0.550 $\pm$ 0.024}\\ 
                  \bottomrule
    \end{tabular}
\end{table*}

\begin{table*}[h!]
    \small
    \centering
    \begin{minipage}[t]{4.5in}
    \caption{Number of samples for each dataset presented in the benchmark.}
    \vspace{0.25cm}
    \label{tab:dataset_num_samples}
    \end{minipage}
    \begin{tabular}{ cccc } 
    \toprule
    \multirow{2}{*}{Task} &   
    \multicolumn{3}{c}{Number of Samples} \\
    \cmidrule(lr){2-4} 
         & Train &  Val & Test  \\ 
        \midrule
        SMP & 103547 & 12943 & 12943 \\
        \midrule
        PIP & 87303 & 31050 & 15268 \\ 
        \midrule
        RES & 3820837 & 192371 & 648372 \\
        \midrule
        MSP & 2864 & 937 & 347 \\
        \midrule
        LBA & 3563 & 448 & 452 \\ 
        \midrule
        LEP & 304 & 110 & 104 \\
        \midrule
        PSR & 25400 & 2800 & 16099 \\ 
        \midrule
        RSR & 12479 & 4000 & 4000 \\ 
        \bottomrule
    \end{tabular}
\end{table*}

\end{document}